\documentclass[journal]{IEEEtran}

\usepackage{amsmath,amsfonts}
\usepackage{algorithmic}
\usepackage{algorithm}
\usepackage{array}
\usepackage[caption=false,font=normalsize,labelfont=sf,textfont=sf]{subfig}
\usepackage{textcomp}
\usepackage{stfloats}
\usepackage{xcolor}
\usepackage{url}
\usepackage{verbatim}
\usepackage{graphicx}
\usepackage{cite}
\UseRawInputEncoding
\usepackage{multirow}
\usepackage{hyperref}
\usepackage{orcidlink}
\begin{document}

\title{Benchmarking Deep Learning Architectures for Urban Vegetation Point Cloud Semantic Segmentation from MLS}

\author{Aditya Aditya\,\orcidlink{0000-0002-7824-8837}, Bharat Lohani\,\orcidlink{0000-0001-8589-192X}, Jagannath Aryal\,\orcidlink{0000-0002-4875-2127},~\IEEEmembership{Member,~IEEE}, Stephan Winter\,\orcidlink{0000-0002-3403-6939},~\IEEEmembership{Senior Member,~IEEE}
\thanks{\textit{(Corresponding Author: Aditya Aditya)}

Aditya Aditya is with the Department of Infrastructure Engineering, University of Melbourne, Parkville, VIC 3010, Australia, and the Department of Civil Engineering, Indian Institute of Technology Kanpur, Kanpur 208016 India. (e-mail: \emph{aditya1@student.unimelb.edu.au}) 

Bharat Lohani is with the Department of Civil Engineering, Indian Institute of Technology Kanpur, Kanpur 208016 India. (e-mail: \emph{blohani@iitk.ac.in}) 

Jagannath Aryal and Stephan Winter are with the Department of Infrastructure Engineering, University of Melbourne, Parkville, VIC 3010, Australia. (e-mail: \emph{jagannath.aryal@unimelb.edu.au; winter@unimelb.edu.au)} }}


\maketitle

\begin{abstract}
\textcolor{red}{This is the preprint version. The paper has been accepted for publication in IEEE Transactions on Geoscience and Remote Sensing. DOI: 10.1109/TGRS.2024.3381976 }\\
IEEE Copyright Notice:
\textcolor{blue}{\textcopyright 2024 IEEE. Personal use of this material is permitted. Permission from IEEE must be obtained for all other uses, in any current or future media, including reprinting/republishing this material for advertising or promotional purposes, creating new collective works, for resale or redistribution to servers or lists, or reuse of any copyrighted component of this work in other works.
}\\
Vegetation is crucial for sustainable and resilient cities providing various ecosystem services and well-being of humans. However, vegetation is under critical stress with rapid urbanization and expanding infrastructure footprints. Consequently, mapping of this vegetation is essential in the urban environment. Recently, deep learning for point cloud semantic segmentation has shown significant progress. Advanced models attempt to obtain state-of-the-art performance on benchmark datasets, comprising multiple classes and representing real world scenarios. However, class specific segmentation with respect to vegetation points has not been explored. Therefore, selection of a deep learning model for vegetation points segmentation is ambiguous. To address this problem, we provide a comprehensive assessment of point-based deep learning models for semantic segmentation of vegetation class. We have selected seven representative point-based models, namely PointCNN, KPConv (omni-supervised), RandLANet, SCFNet, PointNeXt, SPoTr and PointMetaBase. These models are investigated on three different datasets, specifically Chandigarh, Toronto3D and Kerala, which are characterized by diverse nature of vegetation and varying scene complexity combined with changing per-point features and class-wise composition. PointMetaBase and KPConv (omni-supervised) achieve the highest mIoU on the Chandigarh (95.24\%) and Toronto3D datasets (91.26\%), respectively while PointCNN provides the highest mIoU on the Kerala dataset (85.68\%). The paper develops a deeper insight, hitherto not reported, into the working of these models for vegetation segmentation and outlines the ingredients that should be included in a model specifically for vegetation segmentation. This paper is a step towards the development of a novel architecture for vegetation points segmentation.

\end{abstract}


\section{Introduction}
\IEEEPARstart{I}{n} the backdrop of low-cost and compact sensors \cite{luo_indoor_2023}, the deployment of Mobile Laser Scanning (MLS) sensors has seen a steady rise for capturing 3D point clouds in the urban environment. These MLS sensors acquire centimeter-level accurate point clouds with a density up to a few thousand points/m\textsuperscript{2} \cite{wang_survey_2019}. The points are captured from multiple angles and perspectives, providing enhanced scene representation. The point clouds, which are the simplest representation of 3D scenes, can be utilized to extract various objects of interests. Some of the dominant classes in point cloud of urban scenes are buildings, roads, ground, pavements, vegetation, poles and vehicles \cite{hackel_semantic3dnet_2017, roynard_paris-lille-3d_2018, tan_toronto-3d_2020, varney_dales_2020}. Although, each class has its own relevance, vegetation is of notable significance in view of the growing concerns related to sustainable cities and deteriorating urban environment \cite{turner-skoff_benefits_2019, alsalama_mitigation_2021, beecham_using_2020}. Accurate extraction of vegetation points from point cloud, herein called as \emph{vegetation points segmentation}, is crucial for estimation of various metrics related to vegetation. Vegetation points will pave the way for mapping trees and subsequent carbon stock estimation among other contributions \cite{luo_individual_2021, gulcin_assessment_2021}.

In recent years, there has been a significant advancement in deep learning (DL) technology for point cloud segmentation, enabling enhanced spatial scene understanding in the urban environment. DL models are capable of learning rich semantic features necessary for segmentation and demonstrate state-of-the-art results on various point cloud benchmark datasets \cite{hackel_semantic3dnet_2017, roynard_paris-lille-3d_2018, tan_toronto-3d_2020, zhirong_wu_3d_2015, dai2017scannet, 2017arXiv170201105A}. However, models achieving leading results on benchmark datasets may not necessarily achieve similar outcome on datasets collected in a different area, typically on account of varying class-wise composition, point-wise features and the nature of objects within the dataset. In general, a model with higher mean Intersection over Union (mIoU) does not mean that the Intersection over Union (IoU) is higher across all classes. Therefore, overall best model does not guarantee comparable performance to each class \cite{hackel_semantic3dnet_2017}. Accordingly, the selection of a model while dealing with a class specific segmentation is ambiguous. Comprehensive experiments with multiple DL models are required to have some insights in this regard. Furthermore, the employed datasets should incorporate varying scenarios in terms of scene complexity.

As stated above, the selection of DL model for vegetation points segmentation from LiDAR data is ambiguous in the current literature. Consequently, how to select an appropriate model for vegetation points segmentation from amongst several available DL models, remains an unanswered question. Further, there is no understanding, if such a model is invariant to changing scene complexities and per-point features in the dataset. To the problem, we have performed a benchmarking study with seven representative DL models, namely PointCNN, KPConv (omni-supervised), RandLANet, SCFNet, PointNeXt, SPoTr and PointMetaBase. MLS datasets from three study areas are employed, specifically Chandigarh, Toronto3D and Kerala, characterizing multifaceted vegetation along with varying scene complexity, per-point features and class-wise compositions. With terrestrial laser scanning (TLS) supplying data with restricted coverage and airborne laser scanning (ALS) providing data with limited density, MLS datasets have been preferred. 
Faster and rich coverage, adequate point density, and wider availability in urban areas owing to various other applications, have contributed towards the inclusion of MLS datasets. The contributions of our research can be summarized as follows: 
\begin{enumerate}
    \item To the best of our knowledge, this is the first paper to provide benchmarking of representative point-based DL models for vegetation points segmentation from MLS data. We have implemented the models on datasets from three sites for comprehensive assessment.
    
    \item We have developed a correspondence between (a) the physical characteristics of dataset and associated performance, and (b) the architectural designs and functioning behaviour of DL models. 

    \item We have provided specific recommendations towards the selection of a DL model for vegetation points segmentation considering scene complexity and per-point features. Further, we have identified the limitations of DL models while proposing resolutions for better representation and enhanced segmentation accuracy. The suggested components will facilitate a novel architecture for vegetation points segmentation.

\end{enumerate}

\section{Related Works}

Multiple DL architectures have been conceptualized for effective semantic segmentation of point cloud data. There are some projection-based methods \cite{lang_pointpillars_2019, yang_pixor_2018, su_multi-view_2015, boulch_snapnet_2018, han_3d2seqviews_2019} which convert the raw point cloud into images before utilizing 2D CNN for feature learning. Inherent structure and geometrical properties of point cloud are lost while taking projections. Another class of methods are voxel-based \cite{maturana_voxnet_2015, KUMAR201980, meng_vv-net_2019, muzahid_3d_2020} which arrange the point cloud into regular 3D grids called voxels before resorting to 3D CNN for extracting semantic characteristics. These methods are computationally expensive as the number of voxels increase cubically with resolution. 

To resolve these limitations, point-based methods are proposed starting with some seminal work by PointNet \cite{qi_pointnet_2017}. Point-based methods operate on points directly learning per-point features through Multi-layer Perceptron (MLP) or aggregate local features through convolution or utilize a combination of both. Further, some point-based methods are characterized by employing graph convolution or attention mechanism for learning discriminative features. Point-based methods can be classified into the following categories:


\subsection{Point-Wise Deep Learning Based on MLP}
These methods employ MLP to map per-point features to a high-dimensional embedding space. The concept was pioneered by PointNet \cite{qi_pointnet_2017}. It uses shared MLP to learn point-wise features while applying max-pooling to obtain a global feature vector for classification. For segmentation, this is concatenated with intermediate feature maps before employing MLP for dimensionality reduction. Symmetric function and T-Net were used for attaining permutation and translation invariance, respectively. Further, PointNet++ \cite{qi2017pointnet++} was proposed by the same authors to exploit the semantic characteristics of local point features derived from the K-nearest neighbors point set. In PointNet++, the K-nearest neighbors may fall in one orientation. To resolve this, Jiang \textit{et al}. \cite{jiang_pointsift_2018} designed PointSIFT for enhanced local feature representation. PointSIFT module consists of orientation sensitive units capable of encoding scale adaptive information from different orientations. It can be integrated in the PointNet architecture to learn rich semantic features. To extract discriminative contextual features from local neighborhood, Zhao \textit{et al}. \cite{zhao_pointweb_2019} proposed PointWeb. PointWeb consists of Adaptive Feature Adjustment (AFA) module to aggregate features by adaptively learning the impact map signifying element-wise impact indicator between point pairs. To make the best use of local contextual features, some models have been proposed to leverage the geometrical correlation between neighboring points. RandLANet \cite{hu_randla-net_2020} utilizes shared MLP with Local Feature Aggregation (LFA) module to learn per-point semantics. Features are concatenated with their spatial locations while increasing the receptive field for better representation of neighborhood. To resolve the issue of orientation sensitive features, SCFNet \cite{fan_scf-net_2021} was proposed. It uses local polar representation in concatenation with absolute coordinates. Both feature and geometric distances are combined before employing volume ratio for global contextual features. PointNeXt \cite{qian_pointnext_2022} tried to enhance the performance of PointNet++ \cite{qi2017pointnet++} by adopting improved training and scaling strategies. Further, an Inverted Residual MLP (InvResMLP) block was appended with each set abstraction (SA) block to mitigate the vanishing gradient problem. 

\subsection{Point-Wise Deep Learning Based on Convolution}
KPConv \cite{thomas_kpconv_2019} or Kernel Point Convolution applies the convolution operation directly on the input points without any intermediate representation of an MLP. Input points are mapped to uniformly distributed kernel points within a spherical domain. These kernel points learn weights in correspondence to the relative position of kernel points and original point cloud. KPConv can be applied in rigid or deformable configurations. In rigid variant of KPConv, the kernel points are constrained to stay within the sphere at predetermined locations while in deformable version, the location of kernel points is learnable. As part of upsampling, nearest neighbor has been employed. To improve the upsampling process of KPConv, Gong \textit{et al.} \cite{gong_omni-supervised_2021} proposed Receptive Field Component Reasoning (RFCR). It generates target Receptive Field Component Codes (RFCC) to represent existing categories within the receptive field at different layers during the convolution and down-sampling operations. During the decoding stage, the network will reason these RFCCs to have an omni-scale supervision for the predicted semantic labels. To learn rich semantic contextual features, Dense connection-based Kernel Point Network (DenseKPNet) \cite{li_densekpnet_2022} was proposed. It consists of multi-scale KPConv module to extract multilevel features while increasing the receptive field for effective segmentation. Kernel Point Convolution Attention Module (KPCAM) has been added to aggregate local and global semantic features.  

\subsection{Point-Wise Deep Learning Based on a combination of MLP and Convolution}
PointConv \cite{wu2019pointconv} uses MLP and inverse density scale to extend the discretized convolutional filters to continuous point cloud. The network feeds the kernel density estimation into an MLP for computing inverse density and achieves permutation invariance by sharing the weights across all points again by an MLP. SpiderCNN \cite{ferrari_spidercnn_2018} parameterized the convolutional filters as a product of step functions to capture the coarse geometry defined by local geodesic information and Taylor polynomial to express intricate geometric variations. SpiderCNN has been shown to be efficient in association with PointNet. PointCNN \cite{li_pointcnn_2018} employs the \emph{X-Conv} operator, which combines the \emph{X-transformation} with the convolution operation. With \emph{X-Conv}, the input features associated with points are weighed and permuted into a canonical order, followed by the convolution operation on the transformed features.

\subsection{Point-Wise Deep Learning Based on Graph Convolution}
To effectively leverage the rich contextual relationship between points, graph convolutional networks (GCN) \cite{kipf_semi-supervised_2017} are proposed. Landrieu \textit{et al}. \cite{landrieu_large-scale_2018} partitioned the point cloud into geometrically homogeneous elements called superpoint graphs (SPGs) using shape features such as linearity, planarity and scattering. These SPGs provide a rich representation of contextual relationships between object parts and are several orders of magnitude smaller than the original point cloud. Finally, Edge-Conditioned Convolutions, a type of GCN, is utilized for segmentation. To enhance the topological representation of point clouds, EdgeConv \cite{wang_dynamic_2019} was proposed. EdgeConv constructs a local graph called edge features and learns the embeddings, thus describing the relationship between a point and its neighbors. EdgeConv can be incorporated into existing DL pipelines and capable of combining points both in euclidean as well as semantic space. To strengthen the semantic relationship between points, \cite{jiang_hierarchical_2019} proposed edge branch to integrate point features and edge features. To enhance message passing in local regions, edge features are passed to the corresponding point module for integration of contextual information. Du \textit{et al}. 2022 \cite{du_novel_2022} proposed an effective end-to-end graph CNN framework for semantic segmentation referred to as local global graph convolutional method (LGGCM). The key to LGGCM is the local spatial attention convolution (LSA-Conv) which is responsible for the generation of a weighted adjacency matrix for the neighboring points and updation of the nodes by aggregating the features. PointMetaBase \cite{lin_meta_2023} derived useful insights on the building blocks of prevalent models by performing an in-depth analysis. Finally, the learnings were incorporated into a PointMetaBase block. The building blocks were considered with respect to neighbor aggregation, neighbor update, point update and position embedding.

\subsection{Point-Wise Deep Learning Based on Attention Mechanism or Transformers}
Transformer models are readily adopted in point cloud domain after making remarkable strides in natural language processing \cite{vaswani_attention_2017} and image analysis tasks \cite{dosovitskiy_image_2021}. Attention mechanism is at the core of transformer models with its capability to capture long-range dependencies \cite{zhao_point_nodate}. With point clouds being essentially a set embedded irregularly in space with positional attributes, attention mechanism is a natural fit. Point-based transformer models utilize attention mechanism in feature extraction procedures. Point2Sequence \cite{liu_point2sequence_2019} has been conceptualized for exploring local contextual information. Initially, it learns the features of each area scale in a local region before capturing the correlation between these area scales. Hu \textit{et al}. \cite{hu_attention-based_2020} introduced an attention-based local relation learning module for capturing local features. Further, a context aggregation module is proposed with multi-scale supervision for obtaining long-range dependencies between semantically-correlated points. Deng \textit{et al}. \cite{deng_ga-net_2021} proposed an end-to-end global attention network (GA-Net) for learning long-range dependencies in point clouds. To directly process large-scale point clouds, Liu \textit{et al}. \cite{liu_context-aware_2022} proposed context aware network. Local features computed by local feature aggregation module, are combined with long-range dependencies determined by global context aggregation module, to enhance the feature representation. Point Transformer (PT) \cite{zhao_point_nodate} used local self-attention based on vector-attention combined with a trainable and parameterized position encoding mechanism. It integrates the self-attention layer, linear projections and a residual connection, facilitating information exchange between feature vectors. PT is primarily a feature aggregator enabling scalability to large scenes. To improve the scalability and reduce the computational complexity, Self-Positioning point-based transformer (SPoTr) \cite{park_self-positioning_2023} has been proposed. It uses adaptively located self positioning (SP) points for global cross attention considering both local and global shape contexts.


There are only a handful of studies for separating vegetation points from the remaining points using MLS point cloud in the urban setting. Most of the studies are rule-based while being parameters dependent and working within some thresholds. Pu \textit{et al}. \cite{pu_recognizing_2011} partitioned the raw MLS along road directions into ground, on-ground and off-ground points. Later, they used several characteristics of point cloud segments such as shape, size, orientation and topological relationships for segmentation into various categories including vegetation. Zhang \textit{et al}. \cite{zhang_individual_2015} used Normalized Difference Vegetation Index (NDVI) values derived from hyperspectral data in combination with Airborne Laser Scanning (ALS) data to separate vegetation points. Li \textit{et al}. \cite{li_street_2020} used machine learning based detector called Discrete AdaBoost algorithm for segregating vegetation points from MLS point cloud data. Supervised learning was performed by training on statistical features such as width, depth, elevation, density and others, derived from the sphere domain of each point. Although these studies have similar objectives, the approach is heuristic \cite{pu_recognizing_2011}, suffers from lower resolution problem \cite{zhang_individual_2015}, or utilizes hand-crafted features \cite{li_street_2020}. 

To resolve the limitations of heuristic methods, DL based methods have been introduced. Luo \textit{et al}. \cite{luo_individual_2021} have performed such binary semantic segmentation by using the network MS-RRFSegNet \cite{9080553}. However, the selected network was chosen arbitrarily and has been unable to demonstrate dominant results among the selected models. There are some other DL based approaches for tree segmentation but in the natural forest environment using ALS data. Windrim \textit{et al}. \cite{windrim_detection_2020} accomplished tree delineation using faster R-CNN object detector trained on three-channel image generated from the projected point cloud. The image channels are associated with vertical density, maximum height and average return of discretized point cloud. Chen \textit{et al}. \cite{chen_individual_2021} carried out tree segmentation using PointNet based DL framework involving voxelization and crown refinement using height related gradient information. Subsequently, the selection of an appropriate DL model for vegetation points segmentation in the urban setting remains contentious. To resolve the issue, we have performed benchmarking study involving seven representative DL models.      


\section{Materials and Methods}

To comprehensively demonstrate the effectiveness of DL models for vegetation points segmentation, three different MLS datasets have been employed, namely, Chandigarh, Toronto3D \cite{tan_toronto-3d_2020} and Kerala. Toronto3D is an open dataset while the remaining two are developed in-house. Chandigarh and Kerala datasets have been manually labeled into large number of classes and are available at \emph{LiDAVerse.com}. These will be the first of their kind datasets with objects being segmented at instance level. With Chandigarh and Kerala belonging to northern and southern parts of India, respectively, it is pivotal to accommodate the diversity in vegetation. Kerala dataset has the highest point density followed by Chandigarh and Toronto3D datasets in order.


To accommodate the wide range of models, we have selected some representative DL models from each category of point-based methods. Further, the models are selected owing to their strong performance on outdoor point cloud benchmark datasets such as Semantic3D \cite{hackel_semantic3dnet_2017}. The shortlisted models offer adequate potential for vegetation points segmentation and differ considerably in terms of their approach and architectural designs. The selected models are PointCNN, KPConv (omni-supervised), RandLANet, SCFNet, PointNeXt, SPoTr and PointMetaBase. It may be noted that KPConv (omni-supervised) with deformable kernel points, has been implemented which is a modified version of KPConv \cite{thomas_kpconv_2019}. For simplicity, deformable KPConv (omni-supervised) will be herein referred to as KPConv within the scope of this manuscript. Although, RandLANet, SCFNet and PointNeXt belong to a single category but they differ significantly in terms of employed modules. PointNeXt has been selected with its strong emphasis on data augmentation and model scaling strategies. SPoTr and PointMetaBase belong to transformer-based and graph-based networks, respectively. Both are recently conceptualized architectures and have shown superior performance among their peers. PointNeXt-L and PointMetaBase-L have been implemented with regard to the different available versions of each.

\section{Experimental Results and Discussion}

MLS datasets from three different study sites have been employed, namely, Chandigarh, Toronto3D \cite{tan_toronto-3d_2020} and Kerala. 

\subsection{Datasets}

\subsubsection{Chandigarh}
Chandigarh dataset has been acquired by a vehicle mounted MLS system (Riegl VQ-450) in the city of Chandigarh, India by Geokno India Private Limited. The field of view (FOV) was 360$^{\circ}$ while the scan rate and pulse rate were 200 lines/sec and 380 kHz, respectively. The LiDAR range was 330 meters. The data covers a distance of around one kilometer with nearly 195 million points. Each point has five attributes including X, Y, Z, I and class label. These attributes are also referred to as \emph{original per-point features} in this manuscript. The dataset represents a typical well-planned urban environment with low scene complexity. It covers a wide road with regularly planted trees on both sides accompanied by a boundary wall spanning alongside. The dataset has been processed to represent two classes, namely, \emph{vegetation} and \emph{non-vegetation}. A segment of Chandigarh data is shown in Fig. \ref{fig:data}(a). The average proportion of vegetation points out of the total points in the dataset is 37.32\% (Fig. \ref{fig:points}).

\subsubsection{Toronto3D}
\begin{figure}[!h]
\centering
{\includegraphics[width=3.4in]{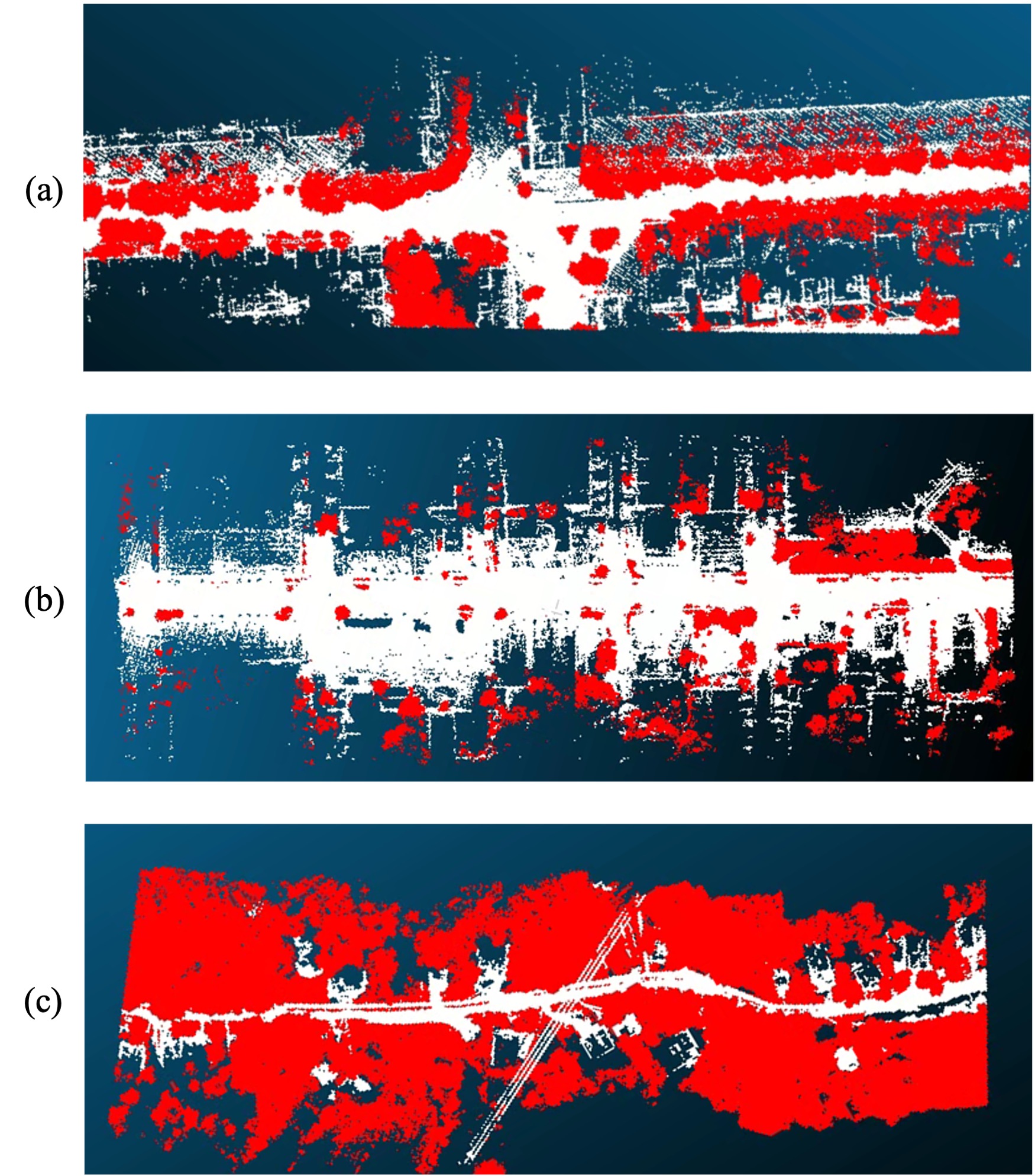}}%
\hfil
\caption{Portion of datasets: (a) Chandigarh (b) Toronto3D (c) Kerala. Red points represent vegetation while white points represent non-vegetation points.}
\label{fig:data}
\end{figure}

\begin{figure*}[!h]
\centering
{\includegraphics[width=6.4in]{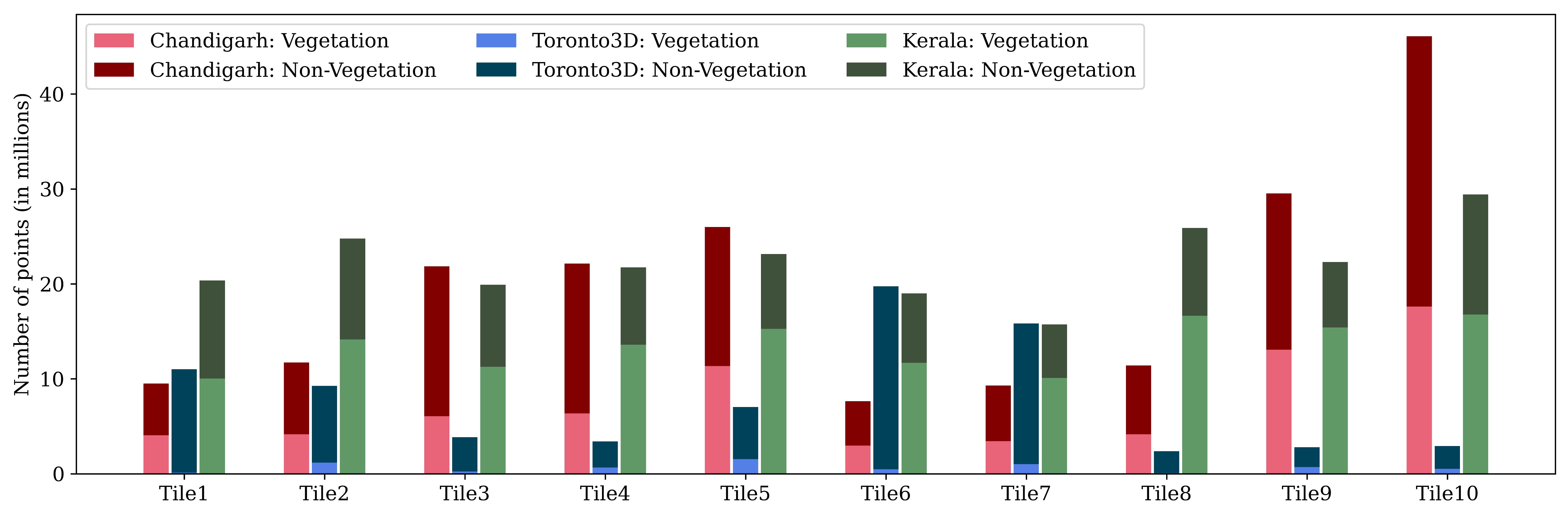}}%
\hfil
\caption{Points distribution across tiles. Chandigarh, Toronto3D and Kerala datasets have been fragmented into 10 tiles for training and testing purposes.}
\label{fig:points}
\end{figure*}

Toronto3D \cite{tan_toronto-3d_2020} is a large scale publicly available point cloud dataset collected by a vehicle mounted MLS system (Teledyne Optech Maverick) in Toronto, Canada. The data represents the urban outdoor environment with low scene complexity. The data consists of around 78 million points collected across a stretch of one kilometer. The data has been segmented into eight classes such as \emph{road, road marking, building, natural, utility line, car, fence and unclassified points}. Each point has eight attributes including X, Y, Z, I, R, G, B and class label.  There is a large variation of point density in the dataset because of the repeated scans at some portions and far range measurements. A segment of Toronto3D data is shown in Fig. \ref{fig:data}(b). The average proportion of vegetation points out of the total points in the dataset is 11.73\% (Fig. \ref{fig:points}).

\subsubsection{Kerala}
Kerala dataset has been captured from Nellanad village situated in Thiruvananthapuram district in the state of Kerala, India with the help of Riegl VQ-450 mounted on an SUV (Tata Xenon) by Geokno India Private Limited. The FOV was 360$^{\circ}$ while the scan rate and pulse rate were 200 lines/sec and 550 kHz, respectively. The LiDAR range was 200 meters. The data has been collected for around one kilometer and consists of about 222 million points. Each point has five attributes including X, Y, Z, I and class label. The dataset represents an environment encompassing a road covered with dense vegetation on both sides with intermittent settlements. The dataset exhibits high scene complexity on account of numerous object categories and multifaceted vegetation. On similar lines to Chandigarh dataset, the data has been processed to represent two classes, namely, \emph{vegetation} and \emph{non-vegetation}. A segment of Kerala data is shown in Fig. \ref{fig:data}(c). The average proportion of vegetation points out of the total points in the dataset is 60.75\% (Fig. \ref{fig:points}).

\subsection{Models}
Seven representative DL models have been selected, at least one from each category of point-based methods for the binary segmentation process. These are PointCNN, KPConv (omni-supervised), RandLANet, SCFNet, PointNeXt, SPoTr and PointMetaBase. 

\subsubsection{PointCNN}
As convolution cannot be directly applied to irregular and unordered point clouds, PointCNN \cite{li_pointcnn_2018} uses an \emph{X-Conv} operator, which is a combination of \emph{X-transformation} and convolution operation. \emph{X-Conv} weighs and permutes the input features associated with the points into a canonical order with \emph{X-transformation}. Further, it applies the convolution operation on the transformed features. \emph{X-Conv} is applied in a hierarchical manner to aggregate the features from the local neighborhood into a set of representative points but each with rich semantic features. Farthest point sampling has been utilized for segmentation as it provides a uniform distribution of sampled points.

\subsubsection{KPConv (omni-supervised)}
Kernel Point Convolution (KPConv) \cite{thomas_kpconv_2019, gong_omni-supervised_2021} performs the convolution operation directly on the input points without any intermediate representation of an MLP. It maps the input points to uniformly distributed kernel points in a spherical domain around each point. Weights are learned by these kernel points along with a correlation factor, which are dependent on the relative positions of kernel points and the original point cloud. Grid subsampling has been used to accommodate varying densities in input data while increasing the cell size in subsequent layers to expand the receptive field. During the decoding stage, the network will reason RFCCs to have an omni-scale supervision for the predicted semantic labels.

\subsubsection{RandLANet}
RandLANet \cite{hu_randla-net_2020} is an efficient and lightweight neural architecture to segment large-scale point clouds. It utilizes random point sampling and a Local Feature Aggregation (LFA) module to infer per-point semantics. LFA consists of a Local Spatial Encoding (LocSE) unit to embed the coordinates so that the features are aware of their relative spatial locations. Further, there is an Attentive Pooling (AP) block to aggregate features and Dilated Residual Block (DRB) to increase the receptive field to preserve geometric details. Initially grid subsampling has been employed followed by random sampling for downsampling procedures. For upsampling, nearest neighbor has been utilized in the decoder part with skip connections.

\subsubsection{SCFNet}
SCFNet \cite{fan_scf-net_2021} is a neural architecture to learn effective features from large-scale point clouds to provide better representation of neighborhood. It employs a Local Polar Representation (LPR) block to furnish a z-axis rotation invariant spatial representation to accommodate orientation sensitive features. In addition, there is a Dual Distance Attentive Pooling (DDAP) block to learn differentiating features considering both the geometric and feature distances. Finally, there is a Global Contextual Feature (GCF) block which takes into account the volume ratio of neighborhood to the global point cloud. To begin with, grid subsampling was employed, followed by random sampling for downsampling operations. As part of upsampling, nearest neighbor has been applied in the decoder part with skip connections.

\subsubsection{PointNeXt}
PointNeXt \cite{qian_pointnext_2022} has performed comprehensive study on training PointNet++ \cite{qi2017pointnet++} with respect to data augmentation, optimization techniques, receptive field and model scaling. It has demonstrated the significance of appropriate training strategies by achieving superior performance from PointNet++ architecture while maintaining high throughput and less parameters with respect to the existing ones. In addition, it has shown that there is no noticeable improvement in accuracy just by increasing the width and depth of the network by using more channels and set abstraction (SA) blocks, respectively while reducing the throughput due to overfitting and vanishing gradients. Further, an Inverted Residual MLP (InvResMLP) block has been designed for efficient model scaling along with some macro-architectural changes like employment of a symmetric decoder and an additional MLP at the beginning. InvResMLP block has been appended with SA block. Different variants of PointNeXt have been proposed based on channel size of stem MLP and number of InvResMLP blocks.

\subsubsection{SPoTr}
Self-Positioning point-based transformer (SPoTr) \cite{park_self-positioning_2023} has been proposed to capture long-range dependencies with reduced complexity. SPoTr consists of two attention modules, namely local points attention (LPA) and self-positioning point-based attention (SPA). LPA learns local shape contexts while SPA performs global cross attention. SPA utilizes a small set of self-positioning (SP) points which are adaptively located to cover the overall shape context. SP points compute attention weights considering both spatial and semantic information through disentangled attention and distributes information to semantically related points in a non-local manner. With these SP points, SPoTr improves the scalability of global attention and captures both local and global shape elements. 


\subsubsection{PointMetaBase}
In PointMetaBase, Lin \textit{et al.} \cite{lin_meta_2023} have provided a detailed analysis of prevalent architectures with respect to neighbor update, neighbor aggregation, point update and position embedding. Max pooling for neighbor aggregation, MLP before neighbor grouping, and explicit position embedding are important considerations enabling better representation with lesser FLOPs. Upon obtaining the necessary insights from the building blocks, PointMetaBase block has been proposed. PointMetaBase has adopted the same scaling strategies and decoder as in case of PointNeXt \cite{qian_pointnext_2022}. PointMetaBase block is appended with PointMetaSA block in each stage.




\subsection{Evaluation Metrics}

Overall Accuracy (OA) (Eq. \ref{equation1}), Intersection over Union (IoU) (Eq. \ref{equation2}) and mean Intersection over Union (mIoU) (Eq. \ref{equation3}) are used to evaluate the performance of models. OA gives the proportion of correctly segmented points out of the total number of points in the dataset \cite{maxwell2021accuracy}. However, it does not takes into account class-imbalance as well as the potential costs associated with incorrect segmentation \cite{alberg_use_2004}. To address the issues, IoU has been used. IoU compensates for varying class frequencies and is indicative of the alignment between the predicted and original point cloud for a particular class \cite{everingham_pascal_2010}. The mIoU is calculated by averaging the IoU values across all classes. These metrics are defined as follows:

\begin{align}
OA &= \frac{TP + TN}{TP + FP + TN + FN} \label{equation1} \\ 
IoU_i &= \frac{TP_i}{TP_i + FP_i + FN_i} \label{equation2} \\ 
mIoU &= \sum_{i=1}^{n} \left( \frac{TP_i}{TP_i + FP_i + FN_i} \label{equation3} \right) / n  
\end{align}
where TP, TN, FP and FN denote the true positives, true negatives, false positives and false negatives, respectively. \textit{i} represents a particular class and \textit{n} is the number of classes which is two in our case.

\subsection{Methodology and Implementation Details}

\begin{figure}[h!]
\centering
{\includegraphics[width=3.4in]{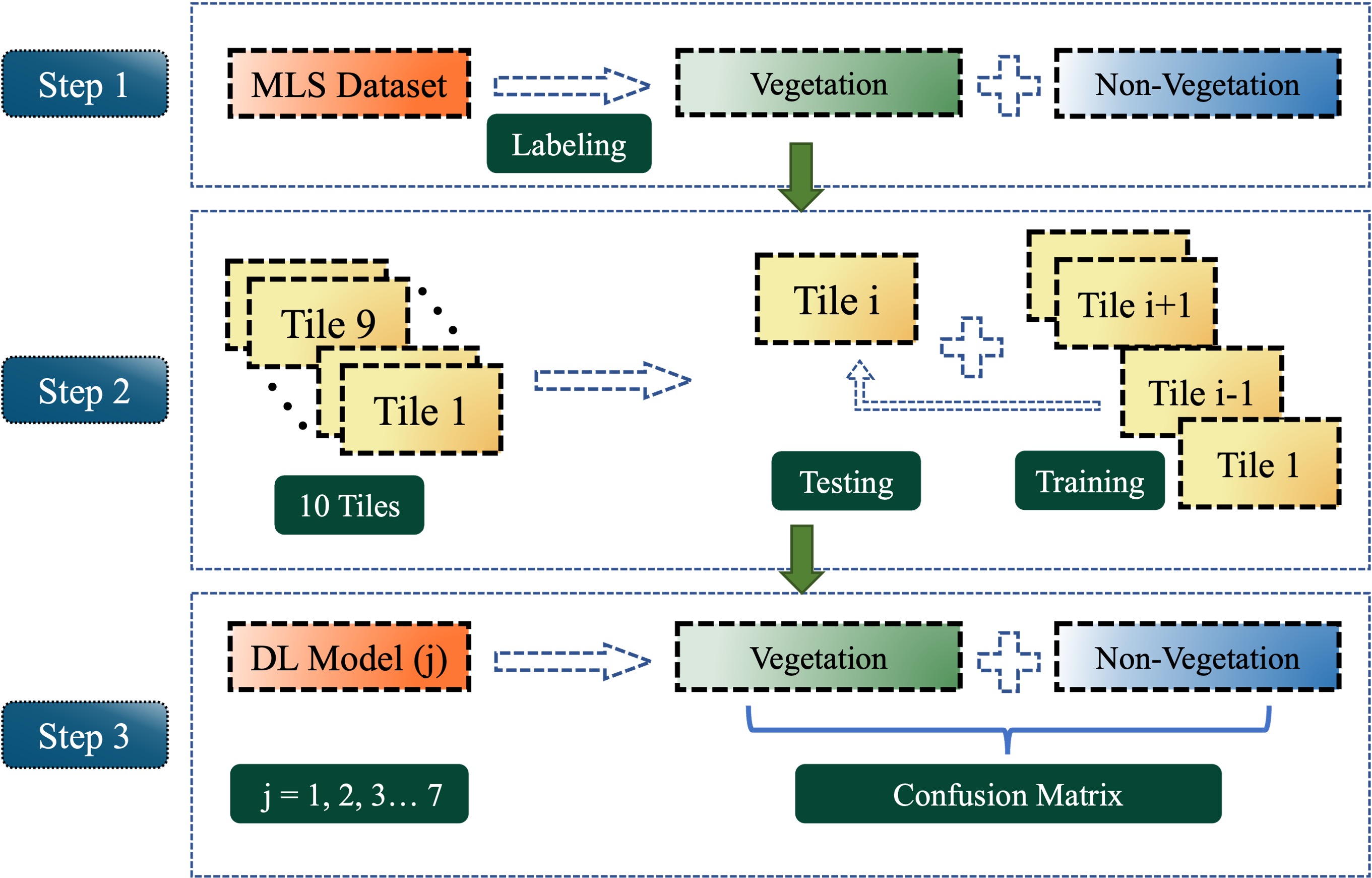}}%
\hfil
\caption{Three step procedure for experimentation. First step is data labeling in two classes. Second step is tiles generation. Final step is model training and testing. \emph{i} can assume integral values between 1 to 10 subject to ten-fold cross validation mode of experiments. Varying \emph{j} is indicative of the seven employed DL models.}
\label{fig:method}
\end{figure}

The experimentation is a three-step procedure (Fig. \ref{fig:method}). In the first step, the available labeled datasets from \emph{LiDAVerse.com} and the Toronto3D dataset, are relabeled to represent only two classes namely, \emph{vegetation} and \emph{non-vegetation}. In the next step, all the three datasets are divided into ten tiles for training and testing purposes. In a single experiment, nine tiles are used for training and validation while the remaining single tile is used for testing. In the final step, the seven DL models are implemented in a ten-fold cross validation mode where every tile has the opportunity to act as a testing tile while the other nine are used for training. The experiments have been performed utilizing a Nvidia Tesla V100 (32 GB) and processor Intel Xeon Gold 5218 2.30 Ghz X 32. PointCNN, KPConv (omni-supervised), RandLANet, and SCFNet are implemented on TensorFlow framework while PointNeXt, SPoTr and PointMetaBase are implemented on PyTorch framework. Further details on implementation are mentioned in Tab. \ref{tab:implement}.  

\begin{table*}[!h]
\fontsize{6pt}{6pt}\selectfont
\centering
\caption{Implementation details of the employed DL models. Initial learning rates are mentioned. MGD stands for Momentum Gradient Descent. *Variable epochs, as PointCNN operates in iterations based on training and testing batches.}
\label{tab:implement}
\setlength\extrarowheight{4pt}
\resizebox{6.5 in}{!}{%
\begin{tabular}{|l|c|c|c|c|c|c|c|} \hline
           Hyperparameters   & PointCNN & KPConv & RandLANet & SCFNet & PointNeXt & SPoTr & PointMetaBase \\ \hline \hline
Epochs        & 50*       & 250    & 50        & 50 & 50 & 50  &  50  \\
Batch Size    & 12         & 16       & 4          & 3   & 4 & 4 & 8   \\
Optimizer     & Adam         & MGD       & Adam          & Adam  & AdamW & AdamW &  AdamW   \\
Parameters (in millions)  & 11         & 14.9       & 1.2          & -   & 7.1 & 66.4 & 2.7   \\ 
Learning Rate & 0.01         & 0.001       & 0.01          & 0.01   & 0.01 & 0.01 & 0.01   \\ \hline
\end{tabular}%
}
\end{table*}

\subsection{Semantic Segmentation Results by Datasets}

Dataset-wise segmentation results (Tab. \ref{tab:performance}) have been presented for assessing the behaviour of the models with changing data characteristics such as scene complexity, vegetation diversity and proportion of vegetation points in the dataset. Scene complexity is described by the number of different object categories occurring along with their configuration in the dataset, and variation in objects within a category as well. At large, scene complexity seems to be one of the most important factors affecting the behaviour of models. Chandigarh and Toronto3D datasets with relatively less complex scenes, are demonstrating better results than those of the Kerala dataset.

\subsubsection{Chandigarh}

In general, highest mIoU values are observed for the Chandigarh dataset (Fig. \ref{fig:miou}). This is expected as Chandigarh is a planned city with vegetation planted almost uniformly along the roads without much species level diversification. Further, the scenes are relatively more organized and structured for the Chandigarh and Toronto3D datasets as compared to the Kerala dataset. For the Chandigarh dataset, the OA values are consistent with mIoU observations (Fig. \ref{fig:oa}).

\subsubsection{Toronto3D}
Second highest mIoU values are observed for the Toronto3D dataset (Fig. \ref{fig:miou}). The data has been collected from a systematic and coordinated urban road environment. However, the declining numbers can be attributed to significantly lower proportion of vegetation points in the Toronto3D dataset. Further, the vegetation points appear in a scattered manner, mostly at a distance from the roads. With regard to OA, leading values are observed for the Toronto3D dataset (Fig. \ref{fig:oa}). Again, this can be attributed to the significantly lower proportion of vegetation points in the dataset. With OA not balancing different class frequencies and being biased towards large classes, inflated results are observed in case of high class-imbalance situations \cite{alberg_use_2004}.

\subsubsection{Kerala}
\begin{figure*}[!h]
\centering
\includegraphics[width=6.8in]{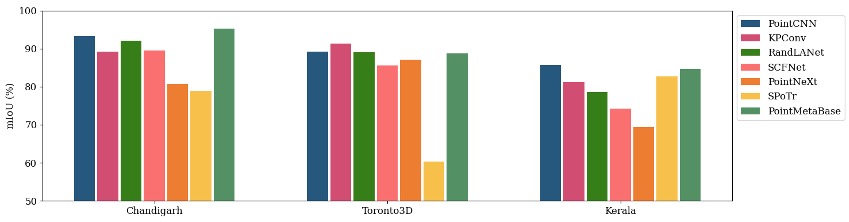}
\caption{Performance of models with respect to mIoU. Highest values are observed on the Chandigarh dataset followed by the Toronto3D and Kerala datasets in order.}
\label{fig:miou}
\end{figure*}

\begin{figure*}[!h]
\centering
\includegraphics[width=6.8in]{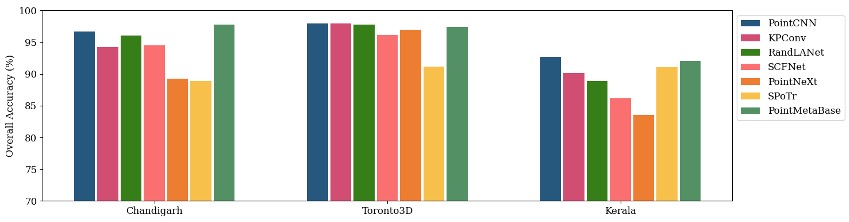}
\caption{Performance of models with respect to overall accuracy (OA). Highest values are observed on the Toronto3D dataset followed by the Chandigarh and Kerala datasets in order.}
\label{fig:oa}
\end{figure*}

Further reduction in mIoU values for the Kerala dataset can be explained by its composition in terms of number of object categories (Fig. \ref{fig:miou}). The dataset exhibits complex environment comprising large number of object categories with high variability within these categories. Further, it is characterized by multifaceted vegetation both in terms of genus and structural characteristics, which are intricately blended. For this dataset, OA values are consistent with mIoU observations (Fig. \ref{fig:oa}).

It may be noted that for a particular dataset, the difference between the mIoU for the models is broadening with increasing scene complexity (Fig. \ref{fig:miou}), again emphasizing the importance of scene complexity towards the performance of models. Chandigarh and Toronto3D datasets, with organized environment, have relatively lower variance as compared to the Kerala dataset with complex scenes.

\begin{table*}[!h]
\fontsize{6pt}{6pt}\selectfont
\centering
\caption{Binary segmentation results: Averaged ten-fold cross validated values. mIou, OA, IoU Veg and IoU N-Veg values are in percentage along with their standard deviation. $\kappa$ denotes the kappa coefficient. Time indicated (approximately) is in hours for combined training and testing.}
\label{tab:performance}
\setlength\extrarowheight{4pt}
\resizebox{6.5in}{!}{%
\begin{tabular}{|l|l|c|c|c|c|c|c|}\hline
                            & Models    & mIoU & OA & IoU Vegetation & IoU  Non-Vegetation & $\kappa$ & Time \\ \hline \hline
\multirow{4}{*}{\rotatebox[origin=c]{90}{Chandigarh}} & 
                              PointCNN     & 93.32 $\pm$ 6.46               & 96.69 $\pm$ 3.39                 & 91.70 $\pm$ 7.82            & 94.93 $\pm$ 5.21                 & 92.86             & 07.0  \\
                            & KPConv       & 89.16 $\pm$ 10.80              & 94.28 $\pm$ 6.08                 & 87.03 $\pm$ 12.38           & 91.28 $\pm$ 9.48                 & 87.87             &   18.0   \\
                            & RandLANet    & 92.08 $\pm$ 6.64               & 96.03 $\pm$ 3.59                 & 90.37 $\pm$ 7.66            & 93.79 $\pm$ 5.86                 & 91.51             &  \textbf{06.0}    \\
                            & SCFNet       & 89.47 $\pm$ 9.55               & 94.52 $\pm$ 5.50                 & 87.33 $\pm$ 10.76           & 91.63 $\pm$ 8.53                 & 88.35             &  06.5    \\
                            & PointNeXt    & 80.80 $\pm$ 14.97              & 89.29 $\pm$ 9.48                 & 77.45 $\pm$ 15.96           & 84.14 $\pm$ 14.67                & 77.94            &  10.5  \\
                            & SPoTr        & 78.88 $\pm$ 14.90              & 88.89 $\pm$ 9.42                 & 73.94 $\pm$ 16.76           & 83.82 $\pm$ 15.62                & 75.31             &  19.5  \\
                            & PointMetaBase& \textbf{95.24 $\pm$ 2.30   }   & \textbf{97.78 $\pm$ 1.03}        & \textbf{93.89 $\pm$ 3.23}   & \textbf{96.59 $\pm$ 1.51}        & \textbf{95.09}    & 10.0  \\

\hline
\multirow{4}{*}{\rotatebox[origin=c]{90}{Toronto3D}}  & 
                              PointCNN      & 89.22 $\pm$ 3.26            & 97.93 $\pm$ 1.35           & 80.87 $\pm$ 6.85           & \textbf{97.56 $\pm$ 1.74}   & 88.05  &   04.0   \\
                            & KPConv        & \textbf{91.26 $\pm$ 5.00}   & \textbf{97.94 $\pm$ 2.14}  & \textbf{84.99 $\pm$ 8.13}   & 97.52 $\pm$ 2.70            &  \textbf{90.43}  &   12.0   \\
                            & RandLANet     & 88.98 $\pm$ 6.59            & 97.78 $\pm$ 1.75           & 80.54 $\pm$ 12.11           & 97.43 $\pm$ 2.01            & 87.44   &   05.0   \\
                            & SCFNet        & 85.61 $\pm$ 12.07           & 96.14 $\pm$ 4.80           & 75.59 $\pm$ 19.52           & 95.62 $\pm$ 5.13            &  82.51 &   \textbf{03.5}   \\ 
                            & PointNeXt     & 87.05 $\pm$ 11.14           & 96.95 $\pm$ 3.03           & 77.58 $\pm$ 19.56           & 96.52 $\pm$ 3.31            & 84.15    &  08.5  \\
                            & SPoTr         & 60.32 $\pm$ 16.36           & 91.17 $\pm$ 8.33           & 29.76 $\pm$ 27.52           & 90.89 $\pm$ 8.31            & 37.88    &  05.5  \\
                            & PointMetaBase & 88.69 $\pm$ 5.74            & 97.37 $\pm$ 2.73           & 80.54 $\pm$ 9.57            & 96.85 $\pm$ 3.41            & 87.33    &  04.0  \\

\hline
\multirow{4}{*}{\rotatebox[origin=c]{90}{Kerala}}     & 
                              PointCNN     & \textbf{85.68 $\pm$ 3.51} & \textbf{92.66 $\pm$ 1.98} & \textbf{88.59 $\pm$ 3.00}   & \textbf{82.77 $\pm$ 4.51}      & \textbf{84.45}  & 08.0  \\
                            & KPConv       & 81.14 $\pm$ 8.22          & 90.18 $\pm$ 4.71          & 85.60 $\pm$ 5.95            & 76.68 $\pm$ 11.04              & 78.75       &   21.0   \\
                            & RandLANet    & 78.64 $\pm$ 7.89          & 88.88 $\pm$ 4.76          & 84.13 $\pm$ 6.81            & 73.15 $\pm$ 9.71               & 75.82       &   04.5   \\
                            & SCFNet       & 74.25 $\pm$ 14.53         & 86.21 $\pm$ 9.20          & 81.32 $\pm$ 8.76            & 67.17 $\pm$ 20.98              & 69.30       &   \textbf{03.5}  \\
                            & PointNeXt    & 69.38 $\pm$ 11.81         & 83.51 $\pm$ 7.12          & 78.21 $\pm$ 8.79            & 60.54 $\pm$ 16.29              & 62.95       &  28.5  \\
                            & SPoTr        & 82.71 $\pm$ 4.83          & 91.11 $\pm$ 2.74          & 86.34 $\pm$ 4.66            & 79.09 $\pm$ 6.39               & 80.85       &  39.0  \\
                            & PointMetaBase& 84.60 $\pm$ 3.34          & 92.05 $\pm$ 2.05          & 87.57 $\pm$ 3.46            & 81.63 $\pm$ 3.92               & 83.25       &  28.5  \\

\hline
\end{tabular}%
}
\end{table*}

\subsection{Semantic Segmentation Results by Models}

\begin{figure}[!h]
\centering
\includegraphics[width=3.45in]{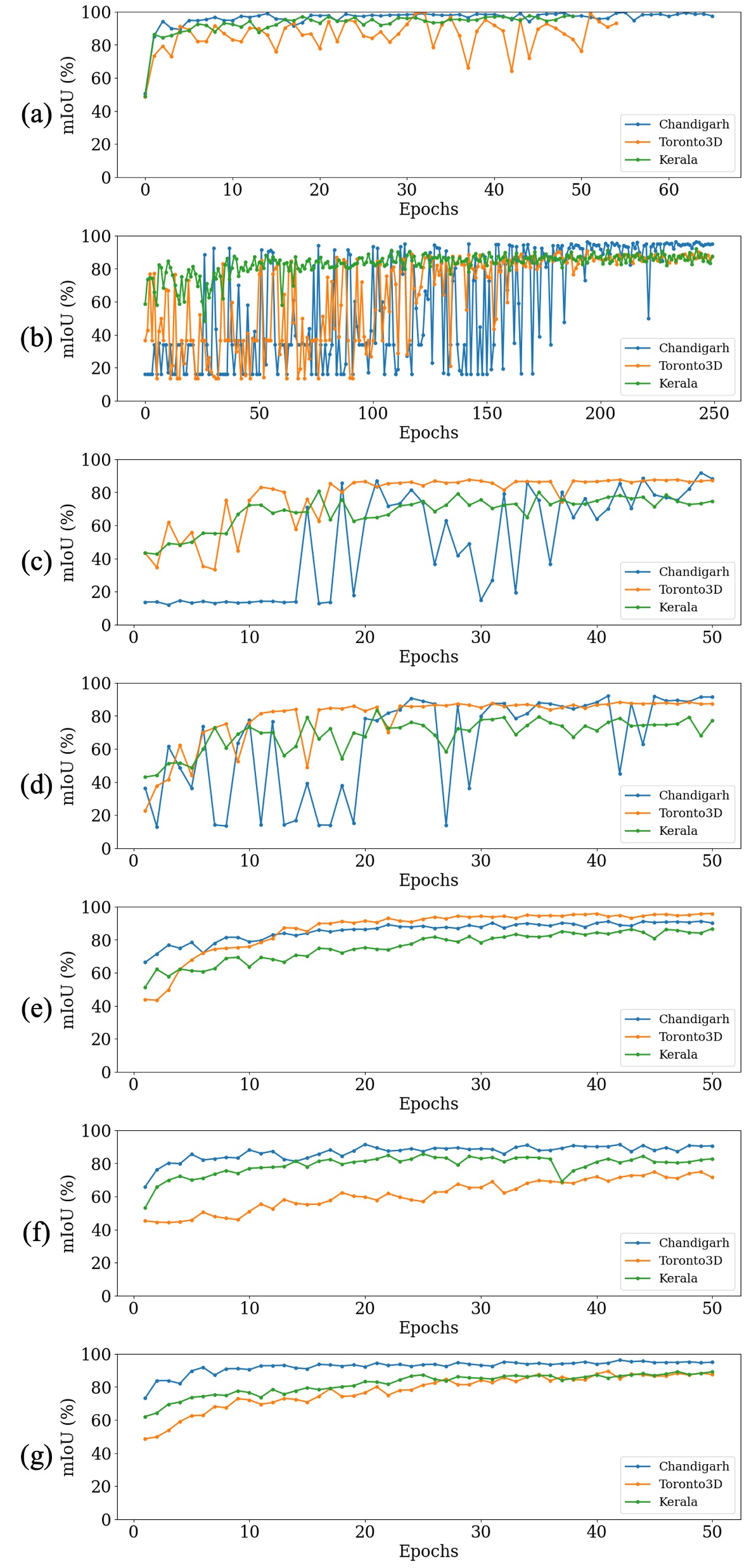}
\caption{Variation of mIoU with epochs: (a) PointCNN, (b) KPConv (omni-supervised), (c) RandLANet, (d) SCFNet, (e) PointNeXt, (f) SPoTr, and (g) PointMetaBase.}
\label{fig:epochs}
\end{figure}

\begin{figure}[!h]
\centering
\includegraphics[width=3.45in]{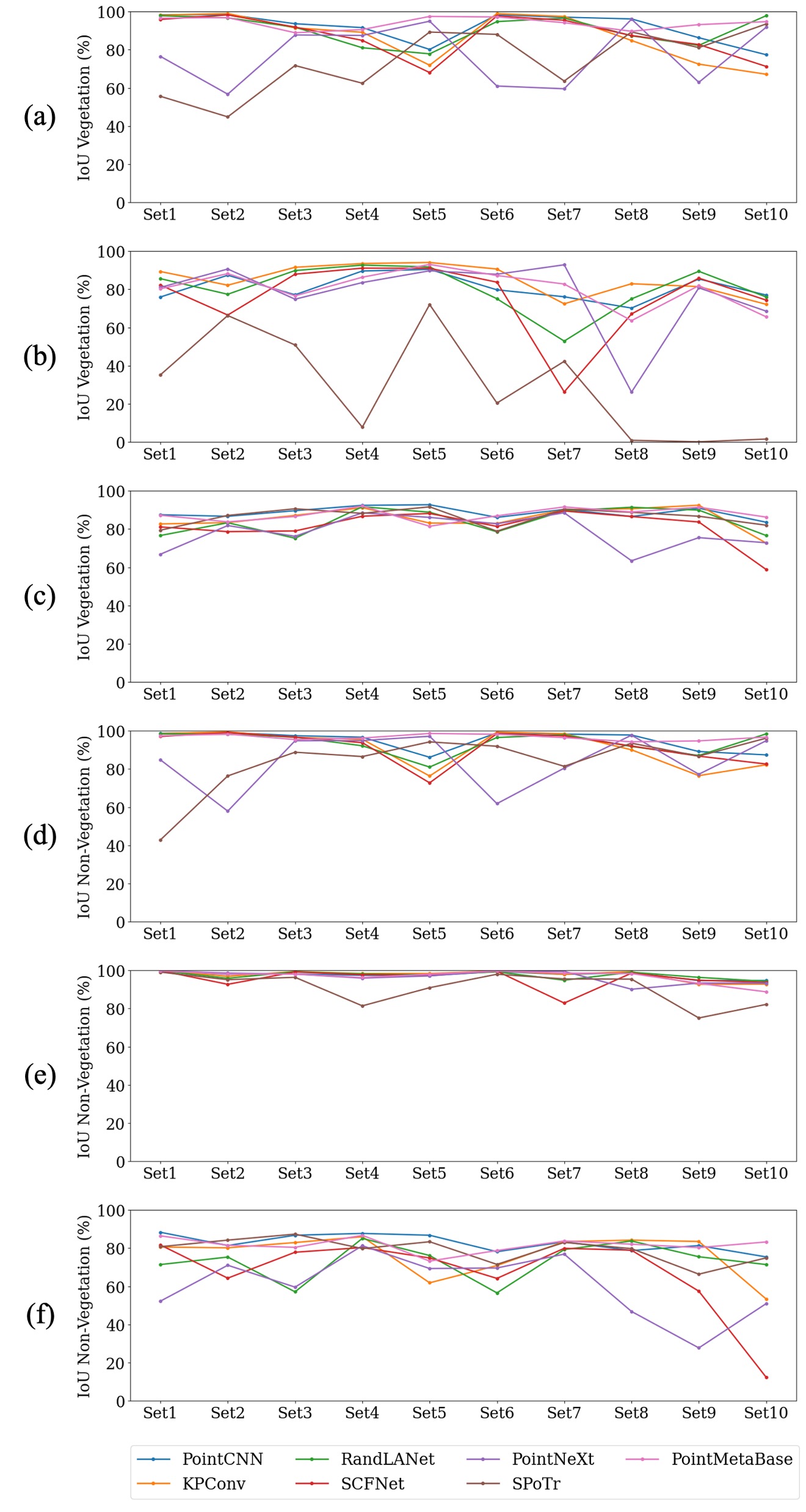}
\caption{Variation of IoU vegetation and IoU non-vegetation with different experimental sets. (a), (b) and (c) indicate IoU vegetation while (d), (e) and (f) show IoU non-vegetation.}
\label{fig:iou}
\end{figure}

\begin{figure}[!h]
\centering
\includegraphics[width=3.45in]{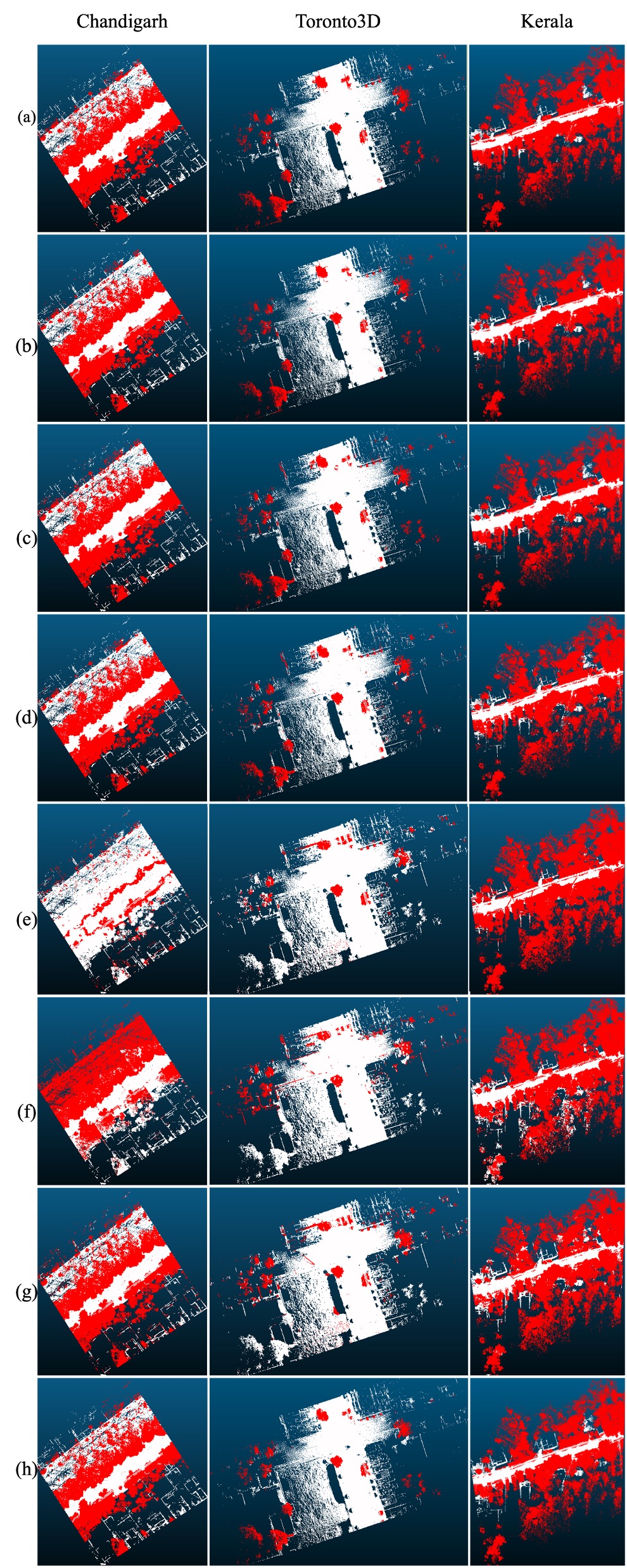}
\caption{Segmentation results on a single tile from each dataset. Red and white points represent vegetation and non-vegetation classes, respectively: (a) PointCNN, (b) KPConv (omni-supervised), (c) RandLANet, (d) SCFNet, (e) PointNeXt, (f) SPoTr, (g) PointMetaBase, and (h) ground truth.}
\label{fig:pointcloud}
\end{figure}

Model-wise segmentation results have been presented for assessing the behaviour of the models with regard to their architectural designs, number of epochs (Fig. \ref{fig:epochs}), time consumption, computational resource utilization and original per-point features in the datasets. Original per-point features are described as initial attributes available with each point in the dataset such as coordinates (X, Y, Z), colors (R, G, B), Intensity (I) and labels. Experimental set wise variation of IoU vegetation and IoU non-vegetation have been shown in Fig. \ref{fig:iou}. Binary segmentation results have been shown in Fig. \ref{fig:pointcloud}.

\subsubsection{PointCNN}
In spite of being conceptualized earliest among the selected models, PointCNN \cite{li_pointcnn_2018} is still furnishing the highest mIoU values on the Kerala dataset. Further, PointCNN has shown second highest mIoU value on the Chandigarh and Toronto3D datasets. This robust performance may be explained by the combination of MLP and convolution, integrated by the network. Initially, the model transforms the point features to a higher dimension with an MLP before using convolution to exploit the spatial relationships between them.

In terms of the number of epochs, the model converges with 50 epochs on all the datasets (Fig. \ref{fig:epochs}). The convergence is explicit for Chandigarh and Kerala datasets. However, there has been some fluctuation in the case of Toronto3D dataset. This can be attributed to lower proportion of vegetation points in the Toronto3D dataset. Five tiles of the Toronto3D dataset have less than seven percent of vegetation points out of total points (Fig. \ref{fig:points}). When the network subsamples a set of 2048 points from each training tile, the proportion of vegetation points is considerably lower. A single batch consists of 12 sets of 2048 points. This proportion varies between batches, resulting in fluctuating behaviour of mIoU with epochs (Fig. \ref{fig:epochs}). Due to the \emph{X-Conv} operator integrating MLP and convolution along with permutation for canonicalization, the model becomes quite complex. Consequently, the network becomes time consuming and computationally intensive (Tab. \ref{tab:implement}).

\subsubsection{KPConv (omni-supervised)}
KPConv \cite{gong_omni-supervised_2021} is furnishing the highest mIoU values on the Toronto3D dataset. This can be attributed to the increased parameters or original per-point features, associated with each point in case of Toronto3D dataset. The kernel-based convolution operation seems to be effective with increasing per-point features. In addition, adopting color drop augmentation significantly boosts the performance \cite{qian_pointnext_2022}. Moreover, increased original per-point features generates effective target RFCCs during the encoding stage, supplementing the performance of the network.

With regard to the number of epochs, convergence is observed with 100 epochs for the Kerala dataset, and 250 epochs for both the Chandigarh and Toronto3D datasets (Fig. \ref{fig:epochs}). However, the model has been operated for 250 epochs for each dataset. This variation in epochs can be explained by the proportion of vegetation points in the dataset. With greater proportion of vegetation points, the possibility of superimposing randomly initialized spherical kernels on vegetation points is higher, leading to faster learning and convergence. However, the convergence is slower as compared to other employed models. In general, larger epochs for convergence, can be described by the convolution operation with spherical kernels comprising deformable points. In a single epoch, the network initiates 5000 spheres and learns features for segmentation \cite{thomas_kpconv_2019}. Further, each epoch is carried out in 500 optimizer steps, making the network time consuming and computationally expensive (Tab. \ref{tab:implement}).

\subsubsection{RandLANet}
Although not showing the best performance, RandLANet \cite{hu_randla-net_2020} can be a preferred model in various scenarios exhibiting less scene complexity, typically in planned cities with organized environment and limited object categories. The model has shown third highest mIoU values on Chandigarh and Toronto3D datasets. Moreover, the mIoU value for the Toronto3D dataset, is comparable to that of PointCNN. Also, the network is effective in scenarios of less vegetative heterogeneity and limited original per-point features. Designed for faster processing of large-scale point clouds, it has shown appreciable outcomes for vegetation points segmentation. The performance seems to enhance with reducing scene complexity while closing the gap with leading performer (Fig. \ref{fig:miou}). Local Feature Aggregation (LFA) module effectively preserves the complex local geometric structures from an expanded neighborhood. 

Concerning the number of epochs, RandLANet seems to converge well with 50 epochs on all datasets (Fig. \ref{fig:epochs}). Further, utilization of random sampling and feed-forward MLP based modules, makes the model lightweight and highly efficient in terms of time and resource consumption (Tab. \ref{tab:implement}).

\subsubsection{SCFNet}
Despite showing dominant performance on the Semantic3D dataset \cite{hackel_semantic3dnet_2017}, SCFNet \cite{fan_scf-net_2021} seems to have issues with the vegetation points segmentation. The model has yielded low mIoU values on all the datasets. This may be associated with the Global Contextual Feature (GCF) block integrated in the architecture. GCF block concatenates the location and volume ratio pertaining to local neighborhood. Volume ratio has been defined as the bounding sphere of the neighborhood to the bounding sphere of the entire point cloud in the global feature context. It considers that this ratio will not change while moving on a particular object but will change at the interface of two objects with changing densities of points. It could be effective to extract the boundaries of objects such as buildings, roads, pavements and poles where the point distribution is uniform for the object. This can be observed from the semantic segmentation results of the network on the Semantic3D dataset \cite{hackel_semantic3dnet_2017}. However, the varying density of points within the vegetation canopy seems to work in a contrary manner. Non-uniform distribution of points within a vegetation canopy will be misinterpreted as different objects particularly by the GCF block, rendering the network ineffective for vegetation points segmentation.

With reference to the number of epochs, the model seems to converge well with 50 epochs on all datasets (Fig. \ref{fig:epochs}). Further, the network is efficient with respect to time and computational resources, owing to the incorporation of random sampling and feed-forward MLP based modules in the architecture (Tab. \ref{tab:implement}).

\subsubsection{PointNeXt}

PointNeXt \cite{qian_pointnext_2022} has achieved competitive accuracy on the Toronto3D dataset while the same has been comparatively lower on Chandigarh and Kerala datasets. This behaviour can be explained by the additional color attributes available in the Toronto3D dataset. Much of the performance of PointNeXt comes from improved training strategies which has a direct relation with the number of attributes. Further, color drop is a strong augmentation technique that significantly improves the segmentation performance when colors are available, as noted and stressed by the authors. Reduced mIoU with respect to other architectures on Chandigarh and Kerala datasets, can be outlined by the employment of only MLP based modules with limited attributes restricting the feature representation capabilities.

The model is showing convergence with 50 epochs (Fig. \ref{fig:epochs}). The network is lightweight in computation and efficient with respect to time (Tab. \ref{tab:implement}).

\subsubsection{SPoTr}

Despite being one of the recent architectures, SPoTr \cite{park_self-positioning_2023} is showing the least mIoU values on Chandigarh and Toronto3D datasets. The mIoU values on Kerala dataset is comparable to other models. This variation in performance across the datasets is related to the self-positioning (SP) points. These SP points are adaptively located across the data to cover various shape contexts and are used for calculating global cross-attention values. These SP points are located considering the latent vector and features of all input points. With points resembling each other are closely located in the latent space, SP points are located within the convex hull of input points.

With high vegetation diversity in Kerala dataset, multiple SP points are initiated resulting in better coverage. Chandigarh and Toronto3D datasets exhibit almost similar kind of vegetation combined with essentially uniform and well-planned urban environment. With this, only a few SP points are initialized which are not enough to capture the entire shape profile. Furthermore, the network is quite complex for binary segmentation with 66 million parameters (Tab. \ref{tab:implement}). With such a complex network, there is a high possibility of overfitting particularly on datasets with less scene complexity. This has also contributed to such low performance on Chandigarh and Toronto3D datasets.

Regarding the number of epochs, the model converges with 50 epochs (Fig. \ref{fig:epochs}). However, the learning is marginal after a few epochs. With the inclusion of local self-attention and global cross-attention combined with channel-wise point attention, the network is quite computationally expensive and time consuming.


\subsubsection{PointMetaBase}

PointMetaBase \cite{lin_meta_2023} is achieving dominant performance on Chandigarh and Kerala datasets while showing the highest mIoU on the Chandigarh dataset. PointMetaBase adopted all the best practices with respect to different building blocks of prevailing DL models. Meticulous integration of explicit position embedding, plain-EPE-Max point update mechanism and non-learnable aggregation function, is responsible for such dominating performance. Employment of trainable and parameterized position encoding is at the core of the network, enabling it to adapt to the local structure. Further, PointMetaBase has adopted the same scaling strategies as in case of PointNeXt, again enhancing the accuracy. With the availability of additional color attributes in Toronto3D dataset, the point update mechanism doesn't seem to blend well with the position embedding module. This resulted in lower performance on the Toronto3D dataset.

Concerning the number of epochs, the models converges well within 50 epochs (Fig. \ref{fig:epochs}). With just 2.7 million parameters, the network is efficient with respect to time and computational complexity (Tab. \ref{tab:implement}). Applying MLP before neighbors grouping resulted into significant reduction in FLOPs. 


\subsection{Towards a Novel Architecture for Vegetation Points Segmentation}

With regard to comprehensive experimentation in ten-fold cross validation mode involving seven representative DL models, the realized mIoU values are appreciable. However, the associated accuracies can be improved by a novel architecture integrating the necessary building blocks from the implemented models. Moreover, there is not a single model which is giving the highest accuracies across all the datasets, further reinforcing the need of a novel architecture. Besides, contrary to the expectations, transformer-based model is unable to demonstrate dominating performance in spite of all necessary computational resources and longer processing times.



An urban environment is typically characterized by varying amounts of vegetation points. However, this variation may reach its extreme for some areas. Proportion of vegetation points could be significantly lower for some portions while the same can be considerably high for other regions. Similarly, the diversity can assume extremes within the vegetation itself. Some areas may exhibit homogeneous vegetation while others may express remarkably distinct vegetation points. A novel architecture should be capable of incorporating all these possible variations, within the vegetation as well as the urban environment. For the novel architecture, the vegetation points not only include points representing trees, but also points belonging to hedges, bushes and undergrowth.

Furthermore, vegetation points are the only naturally occurring objects in the urban setting. Natural objects exhibit distinctive geometrical characteristics to man-made objects which can be assimilated by a novel architecture. Vegetation point clouds exhibit a characteristic arrangement of points within the canopy that can be exploited to separate them from other classes. The laser pulses can penetrate the canopy through gaps, supplying points from the surface as well as from inside the canopy \cite{escola_mobile_2017}. The scheme of points within canopy, will differ with varying species, orientation and location. However, this arrangement of points is fundamental to vegetation and separates it from other classes in an urban setting, thus offering a potential for their effective segmentation. The subsampling techniques employed by the selected DL models such as grid sampling or farthest point sampling \cite{li_pointcnn_2018, gong_omni-supervised_2021, hu_randla-net_2020, fan_scf-net_2021, qian_pointnext_2022} result in uniform distribution of points. This potentially discards the key distinguishing feature of points distribution within the canopy. The dedicated architecture for vegetation points segmentation should employ learnable sampling method \cite{lang_samplenet_2020} that could effectively harness this differentiating pattern. 


On account of architectural design, a network employing both MLP as well as convolution operations is recommended. This will ensure effective learning of rich semantic characteristics, necessary for segmentation. However, the complexity introduced by such a network can be reduced by the manner of implementing MLP and convolution, typically in a sequential fashion. Further, the architecture may be integrated with specialized modules to preserve geometrical characteristics of neighboring points. Moreover, it should have the ability to expand receptive fields in order to consider wider neighborhood. Inspired by \cite{gong_omni-supervised_2021}, the architecture can employ some intelligent mechanism for upsampling procedures, instead of utilizing the commonly used nearest neighbor approach. These components will be crucial to the novel architecture, enabling enhanced segmentation performance as compared to the existing architectures. Essentially, the dedicated architecture has to be concerned with learning effective features for distinguishing vegetation from remaining points regardless of its ability to differentiate other classes.

\section{Conclusion}

Declining vegetation amid the expanding city footprints and deteriorating environment, necessitates their mapping for proper management strategies, particularly in urban setting. At the same time, spatial heterogeneity of urban forests and the diverse structure of trees combined with the complexity in urban areas, pose several challenges. DL for point cloud segmentation has shown significant progress recently, offering state-of-the-art results on various benchmark datasets. These benchmark datasets comprise several classes with DL models aiming to achieve higher mIoU values. However, class specific segmentation with respect to vegetation points has not been explored. Accordingly, the selection of a DL model while dealing with vegetation points segmentation, is ambiguous. To resolve the problem, we have performed a benchmarking study with seven representative point-based DL models, namely PointCNN, KPConv (omni-supervised), RandLANet, SCFNet, PointNeXt, SPoTr and PointMetaBase. To comprehensively demonstrate the effectiveness of models, datasets from three study sites are employed, specifically Chandigarh, Toronto3D and Kerala, characterizing multifaceted vegetation along with varying scene complexity, class-wise composition and per-point features. The experimentation has been performed in a ten-fold cross validation mode. Binary segmentation results have been presented in both dataset-wise and model-wise manner, to gain the necessary insights.

Dataset-wise results indicate that scene complexity is one of the most important factors affecting the behavior of models. Chandigarh and Toronto3D datasets, with more organized and structured scenes, are demonstrating better results than those of the Kerala dataset. With reference to model-wise results, specific recommendations have been provided. For datasets exhibiting complex scenes with limited per-point features, PointMetaBase can be preferred with limited computation capabilities while PointCNN can be adopted with adequate computational resources. KPConv (omni-supervised) can be preferred with increasing per-point features, particularly colors. In connection with organized and less complicated scenes, PointMetaBase can be employed with limited per-point features owing to its effectiveness in terms of time and computational resource consumption. KPConv (omni-supervised) can be adopted with increasing per-point features. Surprisingly SCFNet, demonstrating rich performance on the Semantic3D dataset, has shown low accuracies across all the datasets. The lower mIoU values can be attributed to the GCF block integrated in the network. On similar lines, SPoTr being one of the recent transformer-based model, produced least favorable results on Chandigarh and Toronto3D datasets on account of SP points and complex network.

Detailed analysis into the working of these representative models for vegetation segmentation revealed several flaws, thus offering a potential for enhanced segmentation performance through a novel architecture. Careful assimilation of building blocks from the best performing models on various datasets, combined with necessary feature learning capabilities pertaining to vegetation, may provide better segmentation results. With regard to insights, a dedicated architecture should incorporate the following building blocks for better representation: 

\begin{itemize}
    \item[--] Learnable sampling technique.
    \item[--] Explicit position embedding.
    \item[--] Appropriate augmentation and scaling strategies.
    \item[--] Combination of MLP and convolution operation.
    \item[--] Expanding receptive fields.
    \item[--] Preserve geometrical characteristics of neighbors.
    \item[--] Intelligent mechanism for upsampling procedures.
\end{itemize}

Our investigations further revealed that with vegetation class being fundamentally different from other classes in terms of point cloud representation, the same can be exploited by a novel architecture. With laser pulses penetrating the canopy through gaps, points are acquired from the surface as well as from inside the canopy. This characteristic arrangement of points within the canopy, can be utilized to segregate vegetation points. The future work will involve designing a novel and robust architecture for vegetation points segmentation, by eliminating the limitations of current architectures. The novel architecture will be aimed to deliver state-of-the-art performance on all datasets, irrespective of the location, vegetation diversity, scene-complexity, and associated per-point features. Meticulous integration of the mentioned building blocks will be pivotal to the dedicated architecture. The insights gained from this research will be crucial in such developments.



\section*{Acknowledgments}
We conduct this study as a part of cluster project under the Data Science (DS) Research of Frontier and Futuristic Technologies (FFT) Division of the Department of Science and Technology (DST), Government of India, New Delhi. The authors acknowledge Geokno India Pvt. Ltd. and Chandigarh Administration for providing data from \emph{LiDAVerse.com}. 

\bibliographystyle{ieeetr}
\bibliography{References.bib}

\begin{IEEEbiography}[{\includegraphics[width=1in,height=1.25in,clip,keepaspectratio]{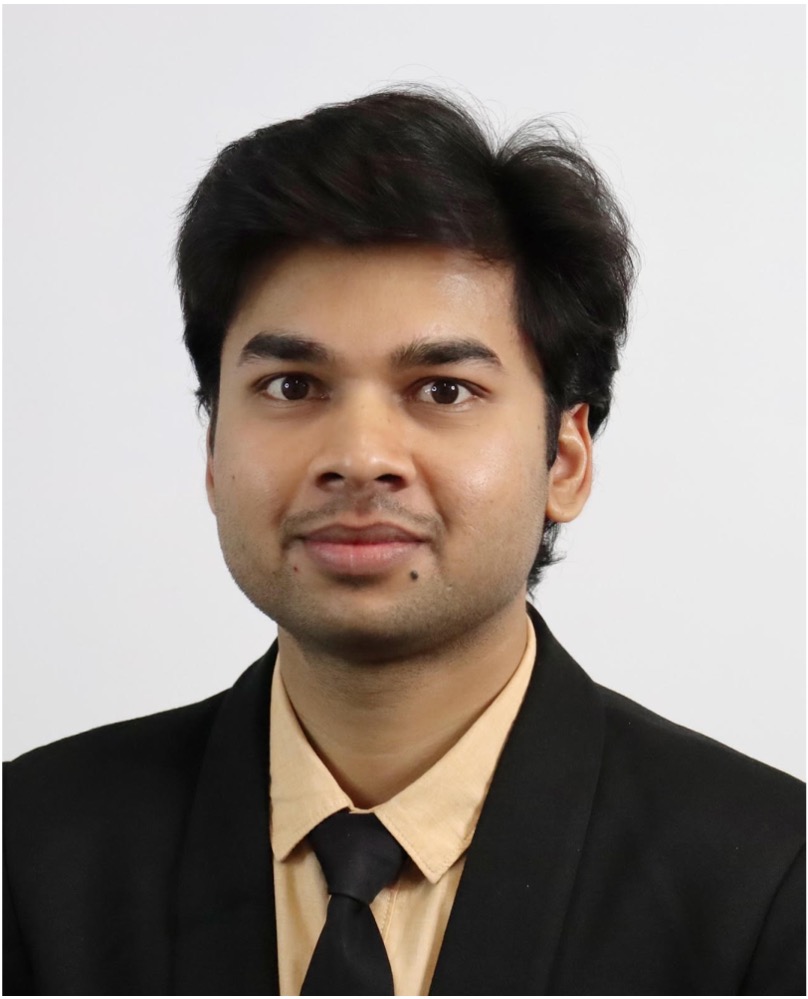}}]{Aditya} received the Bachelor of Technology degree in Mining Engineering from the Indian Institute of Technology (ISM), Dhanbad. 

He is currently a joint PhD candidate at the Department of Infrastructure Engineering, University of Melbourne, Australia, and the Department of Civil Engineering, Indian Institute of Technology Kanpur, India. His research interests include point cloud processing, deep learning for point cloud semantic segmentation and urban forests characterization. 
\end{IEEEbiography}
\vskip -2\baselineskip plus -1fil

\begin{IEEEbiography}[{\includegraphics[width=1in,height=1.25in,clip,keepaspectratio]{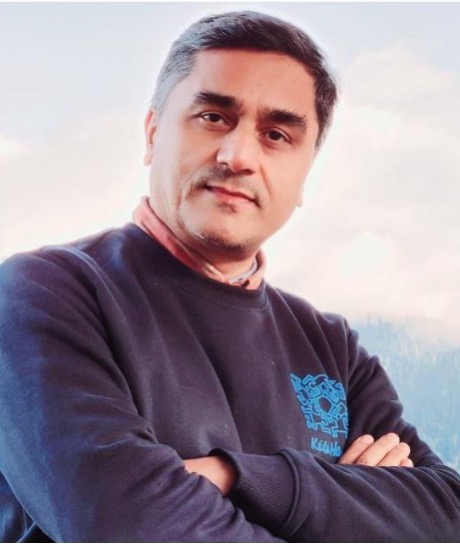}}]{Bharat Lohani} earned his PhD from the University of Reading, the UK, in 1999. He is currently a Professor at the Department of Civil Engineering, Indian Institute of Technology Kanpur, India. 

Dr. Lohani mainly focuses on the modeling of the physical environment using high-resolution remotely sensed data. His current research interests are in LiDAR Data Classification using Deep Learning Techniques, LiDAR application in Forest and Water Conservation, Solar Insolation Estimation, and especially LiDAR simulation for autonomous systems.
\end{IEEEbiography}
\vskip -2\baselineskip plus -1fil

\begin{IEEEbiography}[{\includegraphics[width=1in,height=1.25in,clip,keepaspectratio]{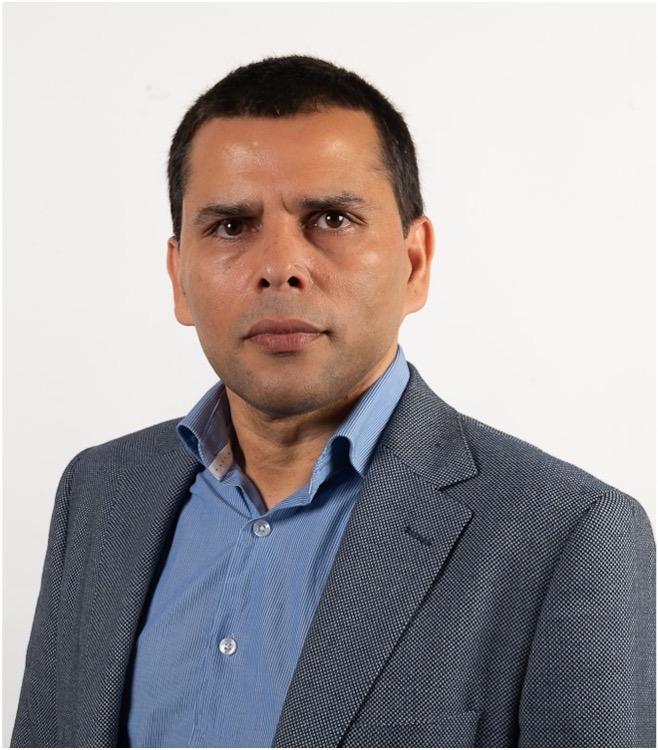}}]{Jagannath Aryal} holds the PhD degree from C-fACS, New Zealand. He is currently an Associate Professor in the Department of Infrastructure Engineering at the University of Melbourne, Australia. His research interests include optimal utilization of Earth observation, geo-information, and geo-statistics to develop new methods for object recognition in urban green space and disaster environment.
Dr. Aryal serves as an Associate Editor for IEEE Transactions on Geoscience and Remote Sensing (TGRS).
\end{IEEEbiography}
\vskip -2\baselineskip plus -1fil

\begin{IEEEbiography}[{\includegraphics[width=1in,height=1.25in,clip,keepaspectratio]{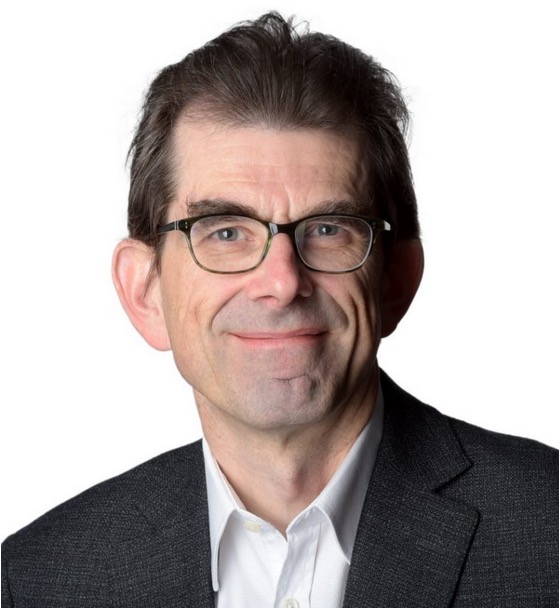}}]{Stephan Winter} holds a PhD from the University of Bonn, and a habilitation from the Technical University Vienna.

He is currently a Professor for Spatial Information Science in the Department of Infrastructure Engineering at the University of Melbourne, Australia. His interest in intelligent mobility, especially for urban sustainability, motivates also data analytics research in urban green spaces.
\end{IEEEbiography}



\end{document}